\documentclass[a4paper,fleqn]{cas-dc}

\usepackage[authoryear]{natbib}
\usepackage{bm}
\usepackage{pbox}
\usepackage{amsmath}
\usepackage{xcolor}

\def\tsc#1{\csdef{#1}{\textsc{\lowercase{#1}}\xspace}}
\tsc{WGM}
\tsc{QE}
\tsc{EP}
\tsc{PMS}
\tsc{BEC}
\tsc{DE}

\begin{document}

\let\WriteBookmarks\relax
\def\floatpagepagefraction{1}
\def\textpagefraction{.001}
\shorttitle{Accepted by Neural Networks}
\shortauthors{Zhihao Zhao et~al.}

\title [mode = title]{Capsule networks with non-iterative cluster routing}

\author[1]{Zhihao Zhao}

\address[1]{Department of Electrical and Computer Engineering, University of Oklahoma, Norman, United states}

\author[1]{Samuel Cheng}
\cormark[1]

\nonumnote{This paper is accepted by Elsevier Neural Networks, Volume 143, November 2021.}
\nonumnote{E-mail addresses: zhihao.zhao@ou.edu(Z.Zhao), \\samuel.cheng@ou.edu (S.Cheng)}
\nonumnote{* Corresponding author.}

\begin{abstract}
	Capsule networks use routing algorithms to flow information between consecutive layers.
		In the existing routing procedures, capsules produce predictions (termed votes) for capsules of the next layer. In a nutshell, the next-layer capsule's input is a weighted sum over all the votes it receives.
		In this paper, we propose non-iterative cluster routing for capsule networks.
		In the proposed cluster routing, capsules produce vote clusters instead of individual votes for next-layer capsules, and each vote cluster sends its centroid to a next-layer capsule.
		Generally speaking, the next-layer capsule's input is a weighted sum over the centroid of each vote cluster it receives.
		The centroid that comes from a cluster with a smaller variance is assigned a larger weight in the weighted sum process.
	Compared with the state-of-the-art capsule networks, the proposed capsule networks achieve the best accuracy on the Fashion-MNIST and SVHN  datasets with fewer parameters, and achieve the best accuracy on the smallNORB and CIFAR-10 datasets with a moderate number of parameters.
	The proposed capsule networks also produce capsules with disentangled representation and generalize well to images captured at novel viewpoints.
	The proposed capsule networks also preserve 2D spatial information of an input image in the capsule channels: 
	if the capsule channels are rotated, the object reconstructed from these channels will be rotated by the same transformation.
	Codes are available at \href{https://github.com/ZHAOZHIHAO/ClusterRouting}{https://github.com/ZHAOZHIHAO/ClusterRouting}.
\end{abstract}


\maketitle

\section{Introduction}
Convolutional neural networks (CNNs) have been very successful in many computer vision tasks, such as image classification \citep{AlexNet, ResNet}, object detection \citep{FasterRcnn, Yolo} and instance segmentation \citep{MaskRcnn}.
However, they sometimes fail to recognize an object captured from novel viewpoints which are not covered in the training data \citep{Engstrom2017, Alcorn2019}.
One purpose of capsule networks is to overcome this problem \citep{capsule2011, Dynamic, Hinton2}. 
Compared with CNNs, capsule networks have the following two major distinctions.
First, the basic unit of capsule networks is a capsule composed of a group of neurons, while the basic unit of CNNs is a single neuron.
A capsule thus can potentially represent multiple properties of an object, such as thickness and scale.
Second, a data-dependent routing procedure is conducted between two consecutive capsule layers, while the flow of information in conventional CNNs is data-independent. 

In the existing routing procedures, capsules produce predictions (termed votes) for the next-layer capsules.
The input of a next-layer capsule is formulated as a weighted sum over all the votes it receives. 
Then its content may be computed from its input by a ``squashing'' function \citep{Dynamic} or by layer normalization \citep{Inverted}.
Iterative routing procedures alternately update the capsule's content and the weights used for formulating the capsule's input through several iterations \citep{Dynamic, Hinton2}.
In contrast, non-iterative routing procedures compute the weights and capsule's content with a straight-through process \citep{Karim2019, Choi2019}. 
By simplifying the iterations to a single forward-pass, non-iterative routing procedures release the computational burden of iterative routing procedures.

We propose non-iterative cluster routing and apply it to capsule networks.
In contrast to the existing routing procedures, in the proposed cluster routing,
capsules produce vote clusters instead of individual votes for capsules of the next layer.
A vote cluster comprises many votes, and each vote may be produced based on a different previous-layer capsule.
A cluster's votes close to each other indicate that the same information is extracted from various previous-layer capsules. 
Thus the vote cluster's variance can be utilized to represent its confidence in the information it encodes.
The input of a next-layer capsule is a weighted sum over the centroid of each vote cluster it receives, and the centroid that comes from a cluster with a smaller variance is assigned a larger weight.
On several classification datasets, capsule networks with the proposed cluster routing achieve the best accuracy compared to the state-of-the-art capsule networks.
Our capsule networks also preserve advantages of the previous types of capsule networks --- producing capsules with disentangled representation \citep{Dynamic, Choi2019} and generalizing well to images captured from novel viewpoints \citep{Hinton2, Karim2019}.
We also show that the proposed capsule networks preserve 2D spatial information such as the rotational orientation of an input image through a reconstruction experiment, where we first rotate the capsule channels by a transformation $T$, then observe if the reconstructed object is rotated by the same transformation $T$.

We outline the contributions of our work as the following:
\begin{itemize}
	\item A novel non-iterative cluster routing is proposed for capsule networks.
	In the proposed cluster routing, capsules produce vote clusters instead of individual votes for next-layer capsules.
	The variance of a vote cluster is utilized to compute its confidence in the information it encodes.
	While computing a next-layer capsule's content, the vote cluster with smaller variance contributes more than other vote clusters.
	
	\item Compared with the state-of-the-art capsule networks, the proposed capsule networks achieve the best accuracy on the fashion-MNIST and SVHN datasets with the fewest parameters.
	On the smallNORB and CIFAR-10 datasets, the proposed capsule networks achieve the best accuracy with a moderate number of parameters.
	
	\item The proposed capsule networks produce capsules with disentangled representation, generalize well to images captured from novel viewpoints, and preserve 2D spatial information of an input image in the capsule channels.
\end{itemize}

\section{Related Works}
\subsection{Capsule networks}
Capsule networks were first introduced by Hinton et al. \citep{capsule2011}.
More recently, they developed capsule networks with dynamic routing \citep{Dynamic} and EM (Expectation-Maximization) routing \citep{Hinton2}.
Capsule networks with dynamic routing yielded disentangled representation of an image; capsule networks with EM routing generalized well to images captured at novel viewpoints.
	However, these routing methods can be improved from the perspective of computational complexity.
	Li et al. \citep{Li2018} approximated the routing procedure with a master branch and an aide branch.  
	Chen et al. \citep{Chen2019} incorporated the routing procedure into the training process. 
	Zhang et al. \citep{FREM} improved the routing efficiency by using weighted kernel density estimation.
	Ahmed et al. \citep{Karim2019} and Choi et al. \citep{Choi2019} computed the coupling coefficients with a straight-through process.
In addition to the works on releasing computational complexity, Ribeiro et al. \citep{VB-Routing} replaced the EM algorithm in EM-routing with Variational Bayes, which improved both the classification accuracy and novel viewpoint generalization.
Tsai et al. \citep{Inverted}  imposed  layer  normalization  as  normalization and replaced the  sequential  iterative  routing  with  concurrent  iterative routing.
Wang et al. \citep{Wang2018} interpreted the routing as an optimization problem that minimizes a combination of clustering-like loss and a Kullback-Leibler regularization term.

Capsule networks were combined with other techniques.
Lenssen et al. \citep{Lenssen2018} used group convolutions to boost the equivariance and invariance of capsule networks. 
Deliege et al. \citep{Deliege2018} embedded capsules in a Hit-or-Miss layer, which resulted in a hybrid data augmentation process and also detected potentially mislabeled images in the training data. 
Jaiswal et al. \citep{Jaiswal2018}, Saqur et al. \citep{Saqur2018} and Upadhyay et al. \citep{Upadhyay2018}
combined capsule networks with generative adversarial networks \citep{Goodfellow2014} to synthesize images.

Capsule networks were also extended to a wide range of applications.
LaLonde and Bagci \citep{LaLonde2018} extended capsule networks to object segmentation by introducing a deconvolutional capsule network. 
Durate et al. \citep{Duarte2018} developed capsule-pooling and applied capsule networks to action segmentation and classification. 
Zhao et al. \citep{Zhao2019} applied capsules to point clouds for 3D shape processing and understanding.
Zhou et al. \citep{Zhou2019} applied capsule networks to visual question answering tasks with an attention mechanism.

\subsection{Attention mechanism}
The routing procedure is close to the attention mechanism of the Transformer \citep{Transformer}, which produces data-dependent attention coefficients that capture the long-range interactions between inputs and outputs.
Some capsule networks adopted the attention mechanism.
Choi et al. \citep{Choi2019} and Karim et al. \citep{Karim2019} proposed attention-based routing procedures that compute the coupling coefficients between capsules without recurrence.
Xinyi et al. \citep{Xinyi2018} used an attention module in a capsule graph network to focus on critical parts of the graphs.

\section{Methods}
\subsection{Capsule networks with dynamic routing}
	In contrast to a traditional neural network composed of artificial neurons, a capsule network comprises capsules. 
	A capsule comprises a group of neurons that jointly represent an object or an object part.
	We present the classic dynamic routing capsule networks \citep{Dynamic} among various types of capsule networks.
	In dynamic routing capsule networks, a capsule is represented as a vector, and the capsule vector's length represents how active the capsule is.
	A capsule $\textbf{u}_{i} \in \mathbb{R}^D$ at the $l$th layer  is transformed to make ``prediction vectors'' $\hat{\textbf{u}}_{j|i}$ for capsules of the $(l+1)$th layer, by multiplying with weight matrices $\textbf{W}_{ij}$,
	\begin{equation}
	\label{prediction}
	\qquad\qquad\;\;\hat{\textbf{u}}_{j|i} = \textbf{W}_{ij}  \textbf{u}_{i},
	\end{equation}
	where $i$ and $j$ are the indices of capsules of the $l$th and $(l+1)$ layer.
	A ``prediction vector'' is also named a vote for the next-layer capsules.
	The input $\textbf{s}_{j}$ to a next-layer capsule is a weighted sum over all votes it receives, as in Eq~\ref{input}.
	The capsule vector $\textbf{v}_{j}$ of a next-layer capsule is ``squashed'' from its input such that the capsule vector's length is between zero and one, as in Eq~\ref{squash}.
	The dynamic routing iteratively updates the weights $c_{ij}$, the weighted sum $\textbf{s}_{j}$ and the next-layer's capsule vector $\textbf{v}_{j}$ by the following equations,
	\begin{equation}
	\label{input}
	\qquad\qquad\;\textbf{s}_{j}^{(t)} = \sum c_{ij}^{(t)}  \hat{\textbf{u}}_{j|i},
	\end{equation}
	\begin{equation}
	\label{squash}
	\qquad\quad\textbf{v}_{j}^{(t)}\;\; = \frac{\| \textbf{s}_{j}^{(t)}\|}{1 + \| \textbf{s}_{j}^{(t)}\|} \cdot \frac{\textbf{s}_{j}^{(t)}}{\| \textbf{s}_{j}^{(t)}\|},
	\end{equation}
	and
	\begin{equation}
	\qquad\;c_{ij}^{(t+1)} = \frac{\exp(b_{ij} + \sum_{r=1}^t \textbf{v}_{j}^{(r)} \cdot \hat{\textbf{u}}_{j|i})}  {\sum_k \exp({b_{ik}} + \sum_{r=1}^t \textbf{v}_{k}^{(r)} \cdot \hat{\textbf{u}}_{k|i})},
	\end{equation}
	where $t$ is the index of iteration, and $b_{ik}$ is the log prior probability.

\begin{figure*}
	\centering
	\includegraphics[width=0.99\linewidth]{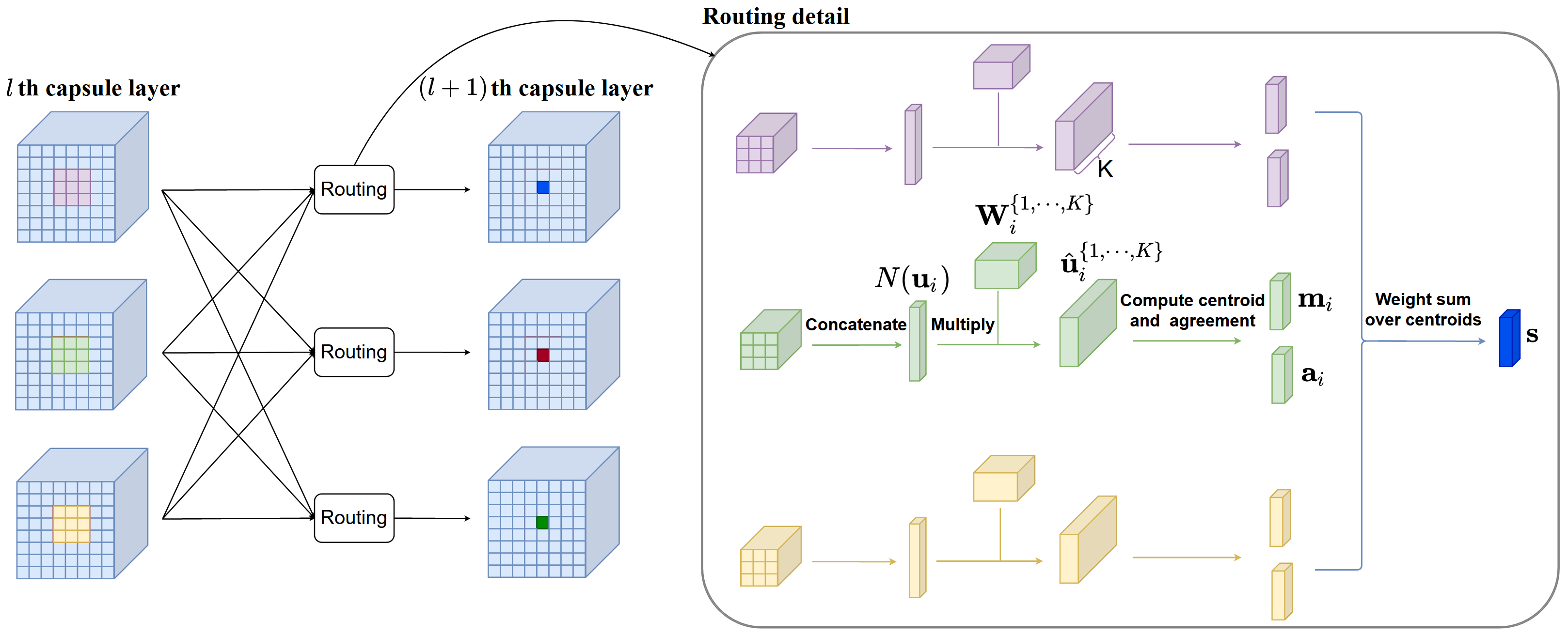}
	\caption
	{Illustration of the proposed cluster routing.
		The left of the figure shows the routing connections between capsules of the $l$th and ($l$+1)th layer; the right of the figure shows the routing detail.
		In the box on the right, $3 \times 3$ capsules of the $l$th layer's $i$th channel are first concatenated into a large capsule $N(\textbf{u}_{i})$, then multiplied with each weight matrix of a weight cluster $\textbf{W}_{i}^{\{1, ..., K\}}$ to produce a vote cluster $\textbf{v}_{i}^{\{1, ..., K\}}$.
		Each vote cluster sends its centroid $\textbf{m}_{i}$ and agreement vector $\textbf{a}_{i}$ to a next-layer capsule of the $(l+1)$th layer.
		The agreement vector is computed based on the variance of the $K$ votes, and is utilized to represent the vote cluster's confidence in its centroid.
		A capsule of the $(l+1)$th layer gets input as a weighted sum over the centroids it receives, and larger weights are assigned to the centroid with larger corresponding agreement vector.
		The content of the $(l+1)$th layer's capsules are computed from their inputs by layer normalization, which is not shown in the figure for the purpose of clarity.
	}
	\label{routing}
\end{figure*}

\subsection{The proposed cluster routing}
	In contrast to the dynamic routing, the proposed cluster routing utilizes vote clusters instead of individual votes.
	A capsule $\textbf{u}_{i} \in \mathbb{R}^D$ at the $l$th layer is multiplied with each weight matrix of a weight cluster $\textbf{W}_{i}^{\{1, \cdots, K\}}$, resulting in a vote cluster $\hat{\textbf{u}}_{i}^{\{1, \cdots, K\}}$ for a capsule at the $(l+1)$th layer. 
	To reduce clutter in the notation, from now on we omit the index $j$ for the next-layer capsules without introducing confusion. Then, for a cluster of weights $\textbf{W}_{i}^{\{1, \cdots, K\}}$, for $k \in \{ 1, \cdots, K\}$, Eq~\ref{prediction} becomes  
	\begin{equation}
	\label{NoNeighbor}
	\qquad\qquad\;\;\hat{\textbf{u}}_{i}^k = \textbf{W}_{i}^k  \textbf{u}_{i}.
	\end{equation}
	Each weight matrix $\textbf{W}_{i}^k$ in a weight cluster may attend to a specific and distinct location of the capsule vector $\textbf{u}_{i}$.
	It is supposed that after the training stage, if all vector locations represent the same object (or object part), each weight matrix will produce the same vote;
	if one vector location does not represent the same object as other vector locations, one or more votes will be different from others.
	Thus the agreement among these votes indicates if the capsule vector correctly represents a certain object.

	Furthermore, we can replace 
	$\textbf{u}_{i}$ in Eq~\ref{NoNeighbor} by its neighborhood to increase the receptive field of each vote.  In practice, 
	we fix with the $3 \times 3$ neighborhood throughout this work and so we replace $\textbf{u}_{i}$
	by the concatenation $N(\textbf{u}_{i}) \in \mathbb{R}^{9D}$ of its neighborhood.
	Then the $k$-th vote in the cluster is produced as follows,
	\begin{equation}
	\label{pre_prediction}
	\qquad\qquad\;\;\;\;\hat{\textbf{u}}_{i}^k = \textbf{W}_{i}^k  N(\textbf{u}_{i}).
	\end{equation}

	A vote cluster, $\hat{\textbf{u}}_{i}^{\{1, \cdots, K\}}$, sends its centroid ${\bf m}_{i} = \frac{1}{K}{\sum_{k=1}^K \hat{\textbf{u}}_{i}^k}$ and an agreement vector ${\bf a}_{i}$ 
	to the next-layer capsule.
	The agreement vector ${\bf a}_{i}$
	is computed by applying the negative log to the 
	votes' standard deviation as follows, 
	\begin{equation}
	\label{sigma}
	{\bf a}_{i} = -\log\left(\sqrt{\frac{1}{K}{\sum_{k=1}^K (\hat{\textbf{u}}_{i}^k - {\bf m}_{i}^k) \circ (\hat{\textbf{u}}_{i}^k - {\bf m}_{i}^k)}}\,\,\right),
	\end{equation}
	where $\circ$ is the Hadamard (element-wise) product. 
	Each of the $C$ capsule channels in the $l$th layer produces a vote cluster for the next-layer capsule.
	Centroids of these vote clusters are weighted summed as follows,
	\begin{equation}
	\qquad\quad\quad\quad\;\; {\bf s} = \sum_{i=1}^C {\bf c}_{i} \circ {\bf m}_{i},
	\end{equation}
	where
	${\bf c}_i = 
	\exp({\bf a}_{i}) \oslash \sum_{j=1}^C\exp({\bf a}_{j})$ and
	$\oslash$ is the Hadamard division.
	We apply layer normalization \citep{layerNormalization} on the weighted sum $\bf s$, resulting in a capsule vector of the next layer as in \citep{Inverted}.

Notice that the matrix product in Eq~\ref{pre_prediction} can be implemented by $D$ ``conv'' filters in popular deep learning libraries such as Tensorflow \citep{Tensorflow} and PyTorch \citep{Pytorch}.
This is also used in Choi et al.'s work \citep{Choi2019}, where the authors name it as convolutional transform.
This decreases the difficulty to program a capsule network with the proposed routing, and also accelerates the running speed because the ``conv'' operation in Tensorflow and Pytorch is highly optimized.

\begin{table*}
	\caption{Comparison on different types of capsule networks.
	}
	\label{RoutingComparison}
	\begin{tabular*}{0.9\textwidth}{ccccc}
		\toprule
		& \pbox{3cm} {Dynamic Routing \\\citep{Dynamic} }& \pbox{3cm} {{EM routing} \\ \citep{Hinton2}} & \pbox{3cm}{Inverted dot- \\product attention \\ routing \citep{Inverted}}  & \pbox{3cm}{The proposed \\ cluster routing}\\
		\midrule
		Routing & sequential iterative
		& sequential iterative  &concurrent iterative  & non-iterative \\
		\midrule
		Poses & vector & matrix &matrix & vector\\
		Activations & n/a (norm of poses) & determined by EM &n/a & n/a\\
		\midrule
		Non-linearity & Squash function & n/a &n/a & n/a\\
		Normalization & n/a & n/a & Layer Normalization & Layer Normalization\\
		\midrule
		Loss Function & Margin loss & Spread loss &Cross Entropy & Cross Entropy\\
		\bottomrule
	\end{tabular*}
\end{table*}

\subsection{Comparisons with related works}
	Although our weight matrix is implemented using convolutional filters, the proposed capsule networks achieve non-linearity by the proposed cluster routing instead of the ReLU activation as in CNNs.
	Table~\ref{RoutingComparison} lists the differences among different types of capsule networks.

	The proposed cluster routing may remind the readers of the group normalization \citep{wu2018group} which also utilizes the mean and standard deviation of a group.
	Group normalization divides the output channels of a convolutional layer into several groups, and normalizes each group with the group's mean and standard deviation, which is similar to other normalization techniques such as batch normalization~\citep{ioffe2015batch}, instance normalization~\citep{ulyanov2016instance} and layer normalization~\citep{layerNormalization}.
	However, in contrast to the proposed cluster routing, the group normalization does not qualify as a routing algorithm, because it has no process similar to the following routing process: i) compute the data-dependent routing weights based on the agreement between votes; ii) compute the input of next-layer capsules as a weighted sum over the votes, where the routing weight are data-dependent.

\section{Experiments}
We evaluate the proposed capsule networks on the following tasks: classification, disentangled representation, generalization to images captured at novel viewpoints, and reconstruction from affine-transformed channels. 
We also visualize the routing weights $\bm{c}_i$, verifying that they are data-dependent as they should be.

\subsection{Classification}
\textbf{Network architectures}
The proposed capsule networks' capacity is related to two hyperparameters, the number of a capsule vector's dimensions, $D$, and the number of a weight cluster's weight matrices, $K$.
We design four variants of the proposed capsule networks by varying $D$ and $K$ while fixing the number of layers as five and the number of channels at each layer as four.
The four variants are named \textit{M-variant1-4} as in Table~\ref{classificationTable}.
During the experiments, we find that the proposed capsule networks also work well even if we use only one capsule channel at each layer.
When using a single channel, we apply $N$ weight clusters on capsules of this single channel which produces $N$ vote clusters for a next-layer capsule.
We also design four variants with a single channel at each layer, named \textit{S-variant1-4} as in Table~\ref{classificationTable}.

Every \textit{M-variant} and \textit{S-variant} has 5 capsule layers, with a stride of 2 at the second and fourth layers.
Each variant is trained for 300 epochs using cross-entropy loss  with stochastic gradient descent. 
The initial learning rate is 0.1 with step decay at every 100 epochs, and the decay rate is 0.1.
A batch size of 64 is used.

\begin{table*}
	\caption{Test error rate comparisons with capsule networks literature and the baseline CNN. ($\cdot$) denotes ensemble size.} 
	\label{classificationTable}
	\footnotesize
	\begin{tabular*}{0.99\textwidth}{lcrcrcrcr}
		\toprule
		\multirow{2}{*}{\textbf{Method}}	& \multicolumn{2}{c}{\textbf{smallNORB}} & \multicolumn{2}{c}{\textbf{Fashion-MNIST}} & \multicolumn{2}{c}{\textbf{SVHN}} & \multicolumn{2}{c}{\textbf{CIFAR-10}} \\
		& Error ($\%$) & Param & Error ($\%$) & Param & Error ($\%$) & Param & Error ($\%$) & Param \\
		\midrule
		\pbox{4cm}{Inverted dot-product attention routing~\scriptsize{\citep{Inverted}}} & \multicolumn{1}{c}{-} & \multicolumn{1}{c}{-} & \multicolumn{1}{c}{-} & \multicolumn{1}{c}{-} & \multicolumn{1}{c}{-} & \multicolumn{1}{c}{-} & 14.83 & 560K\\
		Attention Routing~\scriptsize{\citep{Choi2019}} & \multicolumn{1}{c}{-} & \multicolumn{1}{c}{-} & \multicolumn{1}{c}{-} & \multicolumn{1}{c}{-} & \multicolumn{1}{c}{-} & \multicolumn{1}{c}{-} & 11.39 & 9.6M\\
		STAR-CAPS~\scriptsize{\citep{Karim2019}} & \multicolumn{1}{c}{-} & \multicolumn{1}{c}{-} & \multicolumn{1}{c}{-} & \multicolumn{1}{c}{-} & \multicolumn{1}{c}{-} & \multicolumn{1}{c}{-} & 8.77 & $\simeq$318K\\
		HitNet~\citep{HitNet} & \multicolumn{1}{c}{-} & \multicolumn{1}{c}{-} & 7.7  & $\simeq$8.2M & 5.5 & $\simeq$8.2M & 26.7 & $\simeq$8.2M \\
		DCNet~\scriptsize{\citep{DCNet}} & 5.57 & 11.8M & 5.36 & 11.8M & 4.42 & 11.8M & 17.37 & 11.8M \\
		MS-Caps~\scriptsize{\citep{MS-Caps}} & \multicolumn{1}{c}{-} & \multicolumn{1}{c}{-} & 7.3 & 10.8M & \multicolumn{1}{c}{-} & \multicolumn{1}{c}{-} & 24.3 & 11.2M \\
		Dynamic~\scriptsize{\citep{Dynamic}} & 2.7 & 8.2M & \multicolumn{1}{c}{-} & \multicolumn{1}{c}{-} & 4.3 & $\simeq$1.8M & 10.6 & 8.2M \scriptsize{(7)} \\
		Nair \textit{et al}.~\scriptsize{\citep{Nair}} & \multicolumn{1}{c}{-} & \multicolumn{1}{c}{-} & 10.2 & 8.2M & 8.94 & 8.2M & 32.47 & 8.2M \\
		FRMS~\scriptsize{\citep{FREM}} & 2.6 & 1.2M & 6.0 & 1.2M & \multicolumn{1}{c}{-} & \multicolumn{1}{c}{-} & 15.6 & 1.2M \\
		MaxMin~\scriptsize{\citep{MaxMin}} & \multicolumn{1}{c}{-} & \multicolumn{1}{c}{-} & 7.93 & $\simeq$8.2M & \multicolumn{1}{c}{-} & \multicolumn{1}{c}{-} & 24.08 & $\simeq$8.2M \\
		KernelCaps~\scriptsize{\citep{KernelCaps}} & \multicolumn{1}{c}{-} & \multicolumn{1}{c}{-} & \multicolumn{1}{c}{-} & \multicolumn{1}{c}{-} & 8.6 & $\simeq$8.2M & 22.3 & $\simeq$8.2M \\
		FREM~\scriptsize{\citep{FREM}} & 2.2 & 1.2M & 6.2 & 1.2M & \multicolumn{1}{c}{-} & \multicolumn{1}{c}{-} & 14.3 & 1.2M \\
		EM-Routing~\scriptsize{\citep{Hinton2}} & 1.8 & 310K & \multicolumn{1}{c}{-} & \multicolumn{1}{c}{-} & \multicolumn{1}{c}{-} & \multicolumn{1}{c}{-} & 11.9 & $\simeq$460K \\
		VB-Routing~\scriptsize{\citep{VB-Routing}} & 1.6 & 169K & 5.2 & 172K & 3.9 & 323K & 11.2 & $\simeq$323k \\
		Baseline CNN &  3.76 &  3.30M&  5.21& 3.38M & 3.30 & 3.38M &  7.90 & 3.38M\\
		\midrule
		\textit{S-variant1} (N4K4D13) & 2.80$\pm$0.20 &  150K & 5.19$\pm$0.15 & 152K &  3.89$\pm$0.10 & 156K & 13.33$\pm$0.78 & 156K\\
		\textit{S-variant2} (N4K4D16)& 2.58$\pm$0.32 & 217K &  5.07$\pm$0.13& 215K & 3.77$\pm$0.11 & 219K &  11.58$\pm$0.36 & 219K\\
		\textit{S-variant3} (N8K8D16)& 1.93$\pm$0.22 &  672K& 4.79$\pm$0.13 & 686K & 3.47$\pm$0.07 & 686K &  8.58$\pm$0.15 & 686K\\
		\textit{S-variant4} (N8K8D32)& \textbf{1.57}$\pm$0.13 & 2.53M &  4.68$\pm$0.01&  2.51M  & \textbf{3.37}$\pm$0.03 &  2.55M  &  \textbf{7.37}$\pm$0.06 &  2.55M \\
		\textit{M-variant1} (C4K5D6) & 2.98$\pm$0.24 & 150K & 5.17$\pm$0.07 & 146K & 3.94$\pm$0.07 & 154K & 12.16$\pm$0.30 & 154K\\
		\textit{M-variant2} (C4K5D8) & 3.09$\pm$0.19 & 246K & 5.02$\pm$0.04 & 240K & 3.63$\pm$0.11 & 252K & 11.11$\pm$0.09 & 252K \\ 
		\textit{M-variant3} (C4K8D16) & 1.92$\pm$0.12 & 1.32M &  4.84$\pm$0.07 & 1.30M &  3.56$\pm$0.07 & 1.34M& 8.55$\pm$0.12 & 1.34M\\
		\textit{M-variant4} (C4K8D24)& 1.95$\pm$0.12 & 2.87M& \textbf{4.64}$\pm$0.03& 2.84M & 3.48$\pm$0.14 & 2.89M & 7.89$\pm$0.11 &2.89M \\
		\bottomrule
	\end{tabular*}
	\label{results table}
\end{table*}
\textbf{Datasets and data augmentation} 
	For each dataset, the hyperparameters for data augmentation are tuned by a validation set containing one-fifth of the training images. 
	The models are then retrained with the full training set before testing.
	During the training stage, we add brightness and contrast jitter to an image to perturb its brightness and contrast. 
	For a pixel at position $x$, its value $f(x)$ can be perturbed by $g(x)=\alpha f(x)+\beta$, where $\alpha$ and $\beta$ control contrast and brightness, respectively. 
	In this context, adding random brightness and contrast with a factor of 0.2 to an image means that $\alpha$ is in the range [0.8, 1.2] and $\beta$ is in the range [$-0.2\frac{1}{N_x} \sum_x f(x)$, $0.2\frac{1}{N_x}\sum_x f(x)$], where $N_x$ is the total number of pixels and $\frac{1}{N_x}\sum_x f(x)$ is the mean value of all pixels.

smallNORB \citep{smallnorb} comprises 5 classes of 96 $\times$ 96 stereo images.
The training and test sets both have 24,300  images.
Following the steps in \citep{Hinton2}, we downsample each image to $48 \times 48$ pixels and normalize it to zero mean and unit variance.
During training, we add random brightness and contrast with a factor of 0.2, pad to $56 \times 56$, randomly shift with a factor of 0.2, and randomly cropped to $32 \times 32$.
At test time, we take the center 32 $\times$ 32 crop.

Fashion-MNIST \citep{fashionMnist} comprises 10 classes of 28 $\times$ 28 clothing items. 
The training and test  sets  have  60,000 and  10,000  images,  respectively. 
During training, we add random brightness and contrast with a factor of 0.2, pad to 36 $\times$ 36, take random 32 $\times$ 32 crop, and apply random horizontal flips with probability 0.5. 
At test time, we pad the images to 32 $\times$ 32. 

SVHN \citep{svhn}  comprises 10 digit classes of 32 $\times$ 32 real-world house numbers. 
We trained on the core training set only, consisting of 73,257 images, and tested on the 26,032 images of the test set. 
During training, we add random brightness and contrast with a factor of 0.2, pad to 40 $\times$ 40, and take random 32 $\times$ 32 crop.

CIFAR-10 \citep{cifar10} comprises 10 classes of 32 $\times$ 32 real-world images. 
The training and test sets have 50,000 and 10,000 images, respectively. 
During  training, we add random brightness and contrast with a factor of 0.2, pad to 40 $\times$ 40, take random 32 $\times$ 32 crop, and apply random horizontal flips with  probability 0.5.

ImageNet \citep{imagenet_cvpr09} comprises 1000 classes of real-world images. 
The training and validation sets have 1,281,167 and 100,000 images, respectively.
During training, we resize each image to $256 \times 256$ pixels, take random 224 $\times$ 224 crop and apply random horizontal flips with probability 0.5.
During validation, we take the center 224 $\times$ 224 crop.
Following the previous Ahmed et al.'s work \citep{Karim2019}, the test set is not used.

\textbf{Accuracy comparisons with the state-of-the-arts}
The comparisons between the state-of-the-art capsule networks and the proposed \textit{M-variants} and \textit{S-variants} are listed in Table~\ref{classificationTable}.
On the Fashion-MNIST and SVHN datasets, the proposed capsule networks achieve better accuracy than other types of capsule networks with fewer parameters:
i) on Fashion-MNIST, \textit{M-variant1} achieves an error rate of 5.17\% with 146K parameters, and \textit{S-variant1} achieves an error rate of 5.19\% with 152K parameters;
ii) on SVHN, \textit{M-variant2} achieves an error rate of 3.63\% with 252K parameters, and \textit{S-variant2} achieves an error rate of 3.77\%  with 219K parameters.
For the smallNORB dataset, \textit{S-variant4} achieves the best error rate of 1.57\% with a moderate size 2.53M.
For the CIFAR-10 dataset, \textit{S-variant4} achieves the best error rate of 7.37\% with a moderate size 2.55M; \textit{M-variant4} achieves an error rate of 7.89\% with a moderate size 2.89M.

	\textbf{Classification accuracy on ImageNet}
	For the ImageNet dataset, we design a capsule network variant based on the \textit{M-variant4}.
	Similar to the STAR-CAPS variant designed for ImageNet in \citep{Karim2019}, this variant starts with a 7$\times$7 convolutional layer that outputs 64 channels, followed by a single bottleneck residual block with 256 output channels.
	Then the \textit{M-variant4} is added after the residual block. 
	The Top-1 validation accuracy on ImageNet is 63.87\% and the Top-5 accuracy is 88.98\%, which outperforms the accuracy of 60.07\% and 85.66\% produced by the STAR-CAPS network \citep{Karim2019}.

\textbf{Accuracy comparisons with the baseline CNN}
We compare the variants \textit{M-variant4} and \textit{S-variant4} with a baseline CNN.
The baseline CNN is designed as the following: 
5 ReLU convolutional layers,
layer normalization after the ReLU activation,
256 filters at each layer and 3.38M parameters in total.
As shown in Table~\ref{results table}, the baseline CNN has more parameters, and achieves an higher error rate compared to either of the \textit{M-variant4} and \textit{S-variant4} networks.

\textbf{Experiments on hyperparameters}
For the \textit{M-variants}, we analyze the impact of the hyperparameters $K$ and $D$, while fixing the number of channels at each layer as four.
As shown in Table~\ref{Ablation}, there is a clear trend that both larger $D$ and larger $K$ lead to higher accuracy on the CIFAR-10 dataset.

\textbf{Ablation study}
In the ablation experiment, we train the models from scratch with a constant routing weight $c_i=\frac{1}{C}$, which means the weight becomes data-independent.
As shown in Table ~\ref{AblationNoA}, after removing the data-dependence, the proposed capsule networks' performance drop significantly, which demonstrates the data-dependence is crucial.
\begin{table}
	\caption{Analysis of the hyperparameters $K$ and $D$ on the \textit{M-variants} using the CIFAR-10 dataset. Test error rate and the number of parameters are listed for each setting.}
	\label{Ablation}
	\begin{tabular}{ccccc}
		\toprule
		& D=6 & D=8 & D=16 & D=24\\
		\midrule
		\multirow{2}{*}{K=5} & 12.16\% & 11.11\%  & 9.06\% & 8.06\% \\
		& 154K	  & 252K	 & 872K	  & 1.86M \\
		\midrule
		\multirow{2}{*}{K=8} & 11.29\% & 10.24\%  & 8.55\% & 7.89\% \\
		& 226K	  & 375K	 & 1.34M  & 2.89M \\
		\bottomrule
	\end{tabular}
\end{table}

\begin{table}
	\caption{Ablation study on the proposed capsule network \textit{M-variants}, where each variant's name is abbreviated, e.g., \textit{M-v1}.
		We study the network's classification accuracy when the routing weights $c_i$ are data-dependent or data-independent. 
		The CIFAR-10 dataset is used.}
	\label{AblationNoA}
	
	\begin{tabular}{ccccc}
		\toprule
		Data-dependent routing & \textit{M-v1} & \textit{M-v2} & \textit{M-v3} & \textit{M-v4}\\
		\midrule
		Yes & 12.16 & 11.11 & 8.55  & 7.89\\%
		No  & 35.48 & 33.97 & 33.48 & 32.96\\%
		\bottomrule
	\end{tabular}
\end{table}
	\subsection{Generalization to novel viewpoints}
	We validate the proposed capsule networks' generalization ability to images captured at novel viewpoints using the smallNORB dataset. 
	Following the experiments in \citep{Hinton2},
	we train the proposed capsule networks on one-third of the training data containing azimuths of (300, 320, 340, 0, 20, 40) and test on the test data containing azimuths from 60 to 280;
	for elevation viewpoints, we train on the 3 smaller and test on the 6 larger elevations.
	The validation set consists of images captured at the same viewpoints as in training. 
	For the networks to be compared, we measure their classification accuracy on images captured at novel viewpoints (test set) after matching their classification accuracy on familiar viewpoints (validation set).
	The following networks are compared in Table~\ref{viewpoints}: the baseline CNN model as in~\citep{Hinton2}, EM-routing capsule networks~\citep{Hinton2}, STAR-CAPS networks~\citep{Karim2019}, and capsule networks with the proposed cluster routing.
	The proposed capsule networks achieve the best accuracy of 86.9\% on novel azimuth viewpoints.
	On novel elevation viewpoints, the proposed capsule networks achieve an accuracy of 86.6\%, which is slightly lower than the EM-routing capsule networks while outperforming the baseline CNN.
	\begin{table*}[t]
		\footnotesize
			\caption{A comparison of the smallNORB test error rate on images captured at novel viewpoints when all models are matched on error rate for familiar viewpoints.
				We use the same baseline CNN as in Hinton et al.'s work \citep{Hinton2}.}
			\label{viewpoints}
			\begin{tabular}{lcccc|cccc}
				\toprule  
				& \multicolumn4{c}{\textbf{Azimuth} \ (\%)} & 
				\multicolumn{4}{c}{\textbf{Elevation} \ (\%)} \\
				Model & CNN & EM & STAR-CAPS & Ours & CNN & EM & STAR-CAPS & Ours \\
				$\#$Params & 4.2M & 316K & 318K & 246K & 4.2M & 316K & - & 246K \\
				\midrule 
				Familiar & 96.3 & 96.3 & 96.3 & 96.3 & 95.7 & 95.7 & - & 95.7 \\ 
				Novel    & 80.0 & 86.5 & 86.3 & 86.9    & 82.2 & 87.7 & - & 86.6 \\ 
				\bottomrule
		\end{tabular}
	\end{table*}
	\subsection{Disentangled representation}
\begin{figure*}
		\centering
			\includegraphics[width=\textwidth]{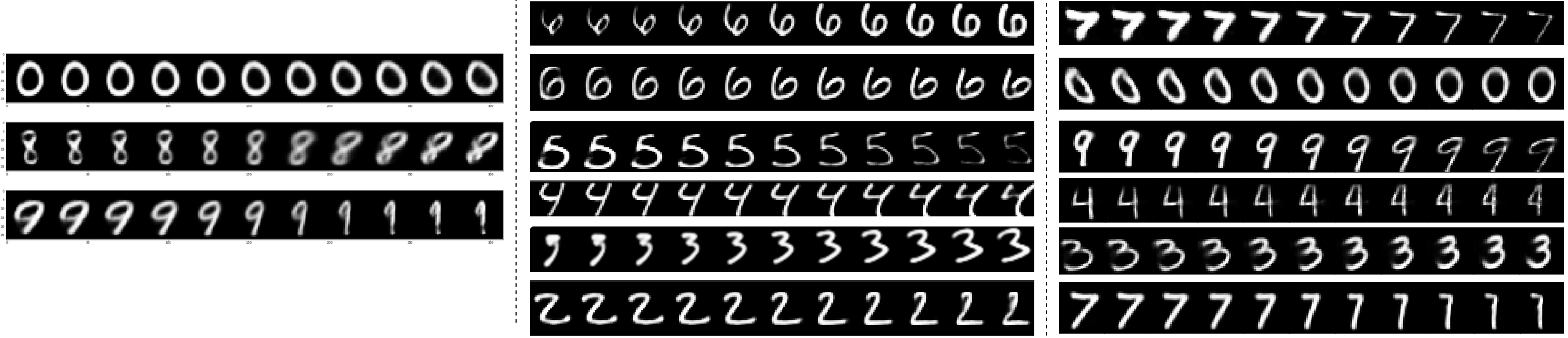}
			\rotatebox{0}{\qquad\quad\quad(a) \parbox{4.7cm}{Attention routing}
				\quad(b) \parbox{5cm}{Dynamic routing } 
				(c) \parbox{5cm}{Proposed cluster routing}}
			\caption{Dimension perturbations on capsules produced by capsule networks with attention routing \citep{Choi2019} (left), dynamic routing \citep{Dynamic} (middle), and the proposed clustering routing (right), respectively. 
				Each row shows the reconstructed images when one dimension of the capsule representing the input digit is tweaked by intervals of 0.05 in the range [-0.25, 0.25]. 
				All three capsule networks produce capsules with disentangled representation -- each dimension of a certain capsule represents a digit's property, such as thickness, skew, and width.}
		\label{disentangled}
\end{figure*}

	Capsule networks with dynamic routing \citep{Dynamic} and capsule networks with attention routing \citep{Choi2019} produce capsules with disentangled representation --- each dimension of a last-layer capsule represents a digit's property, such as thickness, skew, and width.
	The proposed capsule networks also produce capsules with disentangled representation.
	Based on \textit{M-variant2}, we change the number of last-layer capsules to 10, such that each last-layer capsule represents a class of the MNIST dataset.
	For an input image, we mask out all capsule vectors of the last layer except the one representing the input image's class.
	This capsule vector is input to a decoder that reconstructs the input image.
	The decoder has the same architecture as in \citep{Dynamic}, which consists of 3 fully connected layers with 512, 1024, and 784 (784 is the total number of pixels of an MNIST image) neurons, respectively.
	As shown in Figure~\ref{disentangled}, the proposed capsule networks produce capsules with disentangled representation.

\subsection{Reconstruction from affine-transformed channels}
We design a 3-layer capsule network for this task, where the second layer has a stride of 2.
The output channels of the last layer are input into a reconstruction network.
The reconstruction network consists of one upsample layer and two convolutional layers with ReLU and Sigmoid activation.
The Sigmoid convolutional layer outputs the reconstructed image in the range [0, 1].
We transform the last layer's output channels by a transformation $\textit{T}$ and observe the  image reconstructed from the transformed channels.
Ideally, the reconstructed image shall look like the input image transformed by the transformation $\textit{T}$.
The layer normalization is not used due to the simplicity of the MNIST dataset.
The network is trained to perform both classification and reconstruction tasks.

The baseline CNN for this task has a similar architecture to the 3-layer capsule network.
Table~\ref{NumberOfParameters} lists the architecture of the two networks in detail.
A capsule network with dynamic routing \citep{Dynamic} (8.2M parameters)  is also compared.
For the dynamic routing network, we: i) transform the output channels of the second last capsule layer because each channel of the last layer contains only one capsule;
ii) reconstruct the input image using three fully connected layers as in \citep{Dynamic}.

\begin{figure*}
	\centering
	\rotatebox{90}{\quad Proposed 
		\quad\quad\parbox{1.5cm}{Baseline \\ CNN} \quad\quad\parbox{1.5cm}{Dynamic \\ routing}}
	\includegraphics[width=0.8\textwidth]{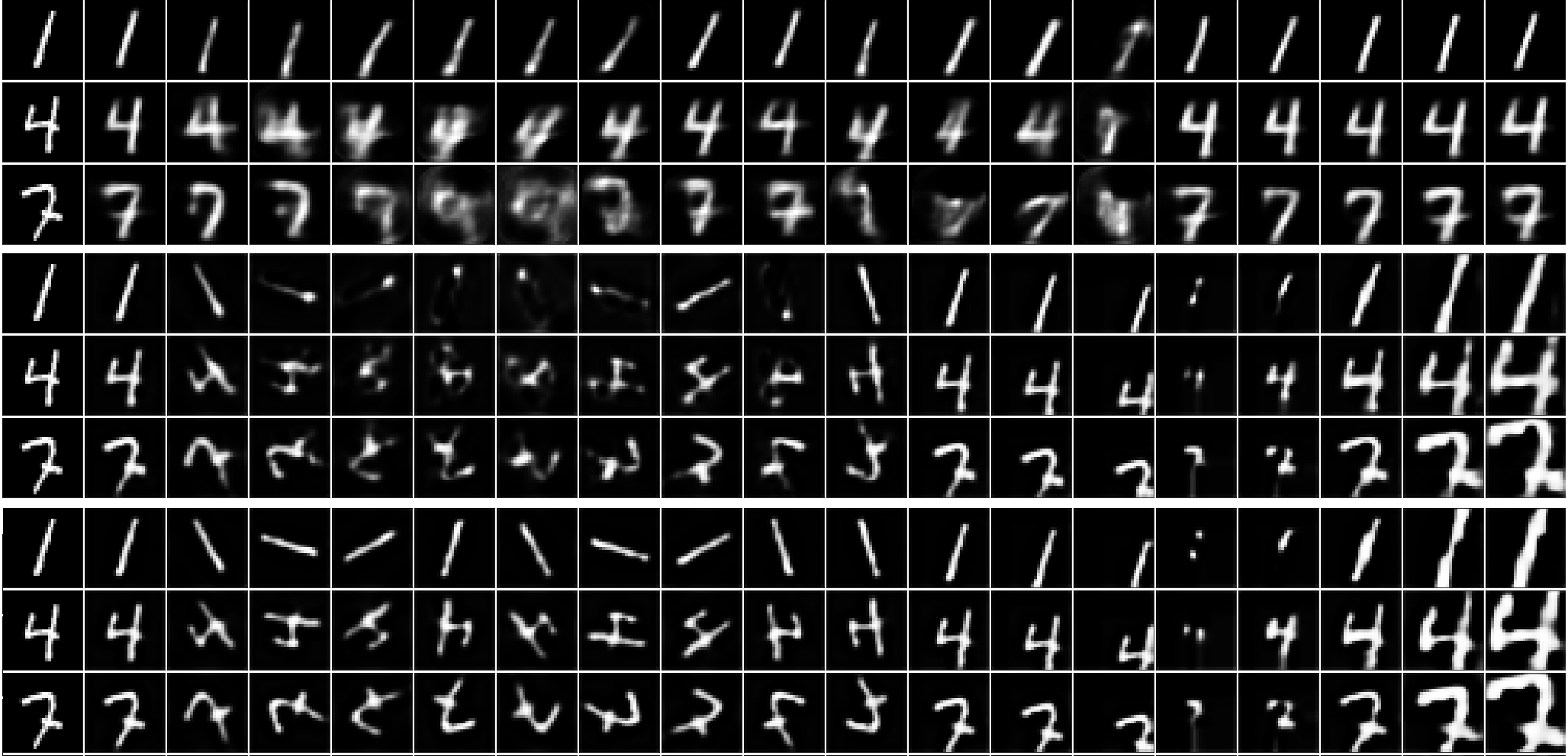}
	\caption{Reconstructed images from capsule channels output by the dynamic routing capsule network \citep{Dynamic} and the proposed capsule network, and reconstructed images from convolutional channels output by the baseline CNN.
		The first column shows groundtruth.
		The other columns show reconstructions from capsule channels (or convolutional channels) applied with the following affine transformations:
		2-9 col: rotation with 0, 45, 90, ..., 315 degrees; 
		10-11 col: horizontal and vertical flip;
		12-14 col: shifting 1, 2, 4 pixels;
		15-19 col: scaling by a factor of 0.5, 0.75, 1.2, 1.5, 2.}
	\label{CapsulecRecons}
\end{figure*}

We show randomly picked reconstructed images in Figure~\ref{CapsulecRecons}. 
A similar pattern for the baseline CNN was observed for about every 4 out of 5 test images, and similar patterns for the dynamic routing and cluster routing capsule networks were observed for every test image.
As shown in Figure~\ref{CapsulecRecons}, the dynamic routing capsule network seems to be always trying to reconstruct the original image.
It produces low-quality reconstructions for large rotations (rotation with a degree from $90^{\circ}$ to $270^{\circ}$), translations, and flips.
The baseline CNN produces fine reconstructions for vertical flips, translations, scaling with a factor larger than 1, and gentle rotations $0^{\circ}$, $45^{\circ}$, $315^{\circ}$ (-$45^{\circ}$),
but fails on large rotations and horizontal flip.
The proposed capsule network produces fine reconstructions for almost all transformations except scaling with a factor less than 1.
In short, the proposed capsule networks succeed in more transformation cases than the baseline CNN and the dynamic routing capsule network.

The quantitative evaluation of reconstructed images is shown in Table~\ref{ReconstructionEvaluate}, using the mean square error (MSE).
Capsule networks with the proposed cluster routing result in the lowest average MSE over the evaluated transformations. 

\begin{table*}[width=\textwidth]
	\caption{Evaluation of reconstructed images using the mean square error.  
		The lower number denotes better performance.
		We evaluate various kinds of affine transformations, including rotations (R-0$^{\circ}$, R-45$^{\circ}$, $\dots$, R-315$^{\circ}$), horizontal and vertical flip (H-flip and V-flip), shifting 1, 2, 4 pixels (Shift-1, Shift-2, and Shift-4), and scaling by a factor of 0.5, 0.75, 1.2, 1.5, 2 (Scale-0.5, Scale-0.75, Scale-1.2, Scale-1.5, and Scale-2).}
	\label{ReconstructionEvaluate}
	
	\begin{tabular}{cccccccccccccccccccc}
		\toprule
		\multirow{2}{*}{Method} & R-0$^{\circ}$ & R-45$^{\circ}$ & R-90$^{\circ}$ & R-135$^{\circ}$ & R-180$^{\circ}$ & R-225$^{\circ}$ & R-270$^{\circ}$ & R-315$^{\circ}$ & H-flip\\
		& V-flip & Shift-1 & Shift-2 & Shift-4 & Scale-0.5 & Scale-0.75 & Scale-1.2 & Scale-1.5 & Scale-2 & Average \\
		\midrule
		\multirow{2}{*}{Dynamic routing} & 0.0176 & 0.0824 & 0.1186 & 0.1025 & 0.1012 & 0.0996 & 0.1197 & 0.0865 & 0.0830 \\& 0.0931 & 0.0488 & 0.0742 & 0.0912 & 0.0896 & 0.0886 & 0.0729 & 0.1712 & 0.2714 & 0.1007\\%
		\midrule
		\multirow{2}{*}{Baseline CNN}  & 0.0025 & 0.0060 & 0.0234 & 0.0310 & 0.0431 & 0.0286 & 0.0223 & 0.0070 & 0.0238 \\& 0.0176 & 0.0632 & 0.1297 & 0.1337 & 0.0238 & 0.0645 & 0.0308 & 0.1096 & 0.0264 & 0.0437\\%
		\midrule
		\multirow{2}{*}{Proposed} & 0.0033 & 0.0047 & 0.0078 & 0.0077 & 0.0092 & 0.0089 & 0.0090 & 0.0053 & 0.0105 \\ & 0.0051 
		& 0.0616 & 0.1279& 0.1332 & 0.0221 & 0.0633 & 0.0329 & 0.1056 & 0.0275 & 0.0359\\
		\bottomrule
	\end{tabular}
\end{table*}

\begin{table}
	\caption{Parameters of a proposed capsule network and a baseline CNN used for reconstructing images from affine-transformed channels.
		Each filter of the baseline CNN is of size $3 \times 3$.
	}
	\label{NumberOfParameters}
	\begin{tabular}{lllcccc}
		\toprule
		& Baseline CNN & Capsule network\\
		&			   & (C4K4D32)		\\
		\midrule
		First layer & 960 (96 filters) & 5,120 \\
		Second layer & 83,040 (96 filters) & 18,944 \\
		Third layer & 13,840 (16 filters) & 18,944 \\
		Linear classifier & 31,370 &  31,370\\
		First recons layer & 4,640 (32 filters) & 4,640\\
		Second recons layer & 289 (1 filter) & 289 \\
		\midrule 
		Total & 134,139 & 79,307\\		
		\bottomrule
	\end{tabular}
\end{table}

\subsection{Analysis of routing weights}
\label{visualization}
A data-dependent routing means that the routing weights $\bf{c}_i$ are dependent on the input image's visual content, unlike the weights matrices that are the same for any input.
However, the reader may come up with a degenerate case where the proposed cluster routing may produce data-independent routing weights:
suppose the weight matrices of a weight cluster are identical or very close to each other, the votes produced by this weight cluster will be identical or very close;
then the routing weight for this vote cluster's centroid will always be almost 1, regardless of the input image's visual content.
To examine if this degeneration happens, we visualize the routing weights.

\begin{figure*}
	\includegraphics[width=\textwidth]{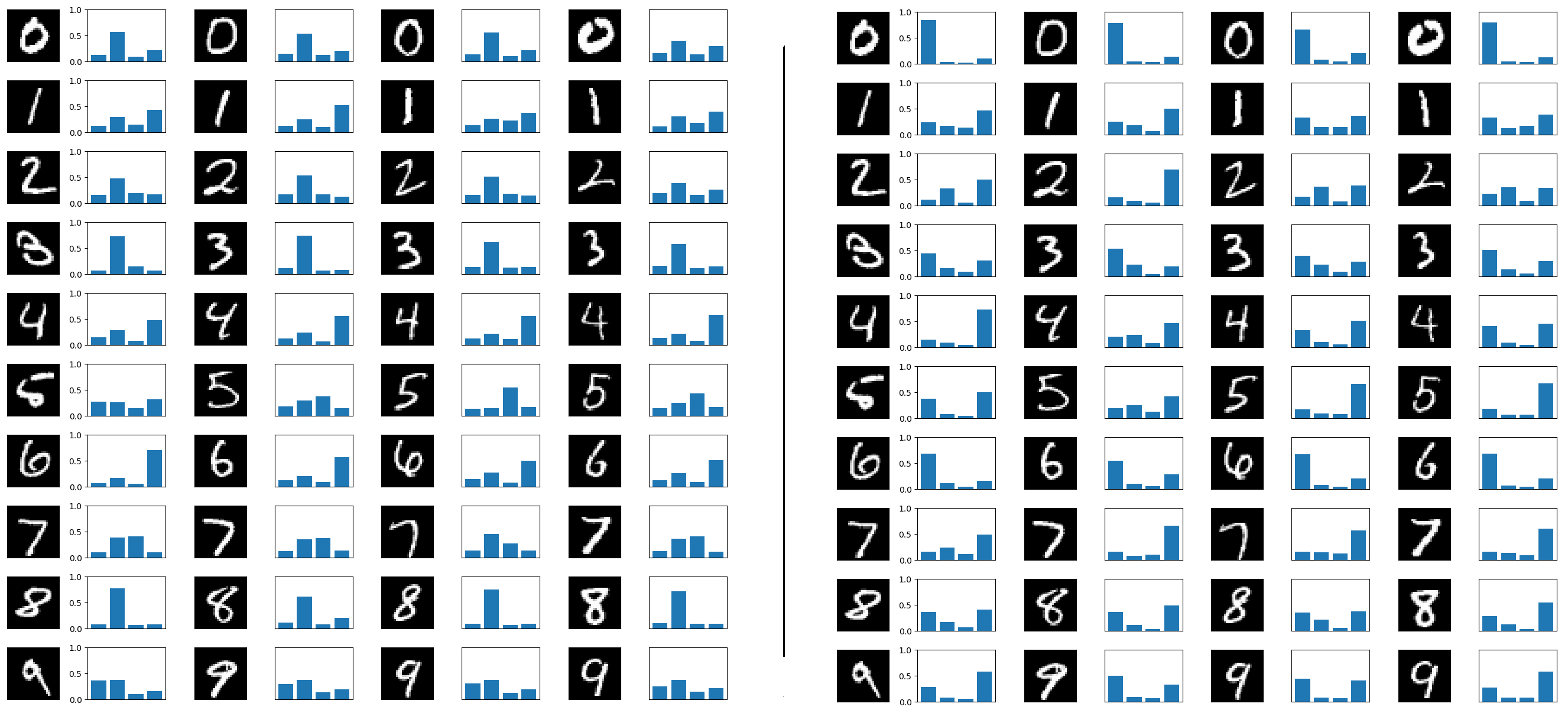}
	\caption{Visualization of routing weights used for last-layer capsules.
		Four bars show routing weights for the four vote clusters that a last-layer capsule receives.
		The left figure shows routing weights for the 8th dimension of the first channel's capsule;
		the right figure shows routing weights for the 2nd dimension of the second channel's capsule.
		Each channel of the last layer is designed to contain only a single capsule.
	}
	\label{RoutingCoefficients}
\end{figure*}

We decide the stride and padding at each capsule layer such that each channel of the last layer contains only one capsule, then visualize routing weights for the last-layer capsules.
Figure~\ref{RoutingCoefficients} shows the routing weights for the four vote clusters that a last-layer capsule receives.
It can be seen from Figure~\ref{RoutingCoefficients} that the proposed capsule networks produce routing weights of the same distribution for images from the same class, but routing weights of different distributions for images from different classes.
This demonstrates that the degenerate case does not happen --- the proposed capsule networks use data-dependent routing weights.

\section{Conclusion}
We propose a non-iterative cluster routing algorithm for capsule networks.
The proposed cluster routing adopts vote clusters instead of individual votes, and the variance of a vote cluster is used to compute its confidence in the information it encodes.
A capsule vector is computed from the vote clusters it receives, where the vote cluster with larger confidence contributes more than other vote clusters.
The experiments show that capsule networks with the proposed cluster routing achieve competitive performance on tasks including classification, disentangled representation, generalization to images obtained from novel viewpoints, and reconstructing images from affine-transformed channels.
In the future, it will be interesting to explore whether some of the vote clusters can be pruned without affecting the performance.

\bibliographystyle{cas-model2-names}
%
\bibliography{capsule}

\end{document}